\documentclass{article}

\usepackage[utf8]{inputenc} 
\usepackage[T1]{fontenc}    
\usepackage{url}            
\usepackage{booktabs}       
\usepackage{amsfonts}       
\usepackage{nicefrac}       
\usepackage{microtype}      
\usepackage{xfrac}

\usepackage{mathtools}
\usepackage{natbib}
\usepackage{xcolor}
\usepackage{url}
\usepackage{amsmath}
\usepackage{amssymb}
\usepackage{graphicx}
\usepackage{subcaption}
\usepackage{nicefrac}
\usepackage{appendix}
\usepackage{enumitem}
\definecolor{darker}{rgb}{0,0.15,0.7}
\usepackage[colorlinks, urlcolor=darker, citecolor=darker, linkcolor=darker]{hyperref} 
\setcitestyle{round}
\usepackage{multirow}


\usepackage{algorithm}
\usepackage{algorithmic}

\usepackage{amsthm}

\begingroup
    \makeatletter
    \@for\theoremstyle:=definition,remark,plain\do{%
        \expandafter\g@addto@macro\csname th@\theoremstyle\endcsname{%
            \addtolength\thm@preskip\parskip
            }%
        }
\endgroup

\newcommand{\vv}[1]{\boldsymbol{#1}}

\usepackage[accepted]{icml2020}

\icmltitlerunning{Non-Autoregressive Neural Text-to-Speech}

\begin{document}

\twocolumn[
\icmltitle{Non-Autoregressive Neural Text-to-Speech}




\begin{icmlauthorlist}
\icmlauthor{Kainan Peng$^*$}{to}
\icmlauthor{Wei Ping$^*$}{to}
\icmlauthor{Zhao Song$^*$}{to}
\icmlauthor{Kexin Zhao$^*$}{to}
\end{icmlauthorlist}
\icmlaffiliation{to}{Baidu Research, 1195 Bordeaux Dr, Sunnyvale, CA.~\ 
 Speech samples can be found in: \url{https://parallel-neural-tts-demo.github.io/}}
\icmlcorrespondingauthor{Wei Ping}{weiping.thu@gmail.com}

\icmlkeywords{Machine Learning, ICML}

\vskip 0.3in
]



\printAffiliationsAndNotice{\icmlEqualContribution.} 

\begin{abstract}
In this work, we propose ParaNet, a non-autoregressive \emph{seq2seq} model that converts text to spectrogram.
It is fully convolutional  and brings  $46.7$ times speed-up over the lightweight Deep~Voice~3 at synthesis, while obtaining reasonably good speech quality.
ParaNet also produces stable alignment between text and speech on the challenging test sentences by iteratively improving the attention in a layer-by-layer manner.
Furthermore, we build the parallel text-to-speech system and test various parallel neural vocoders, which can synthesize speech from text through a single feed-forward pass.  We also explore a novel VAE-based approach to train the \emph{inverse autoregressive flow}~(IAF) based parallel vocoder from scratch, which avoids the need for distillation from a separately trained WaveNet as previous work.
\end{abstract}

\section{Introduction}
\label{introduction}
%
Text-to-speech~(TTS), also called speech synthesis, has long been a vital tool in a variety of applications, such as human-computer interactions, virtual assistant, and content creation.
Traditional TTS systems are based on multi-stage
hand-engineered pipelines~\citep{taylor2009tts_book}.
In recent years, deep neural networks based autoregressive models have attained state-of-the-art results, including high-fidelity audio synthesis~\citep{oord2016wavenet}, and much simpler \emph{seq2seq} pipelines~\citep{sotelo2017char2wav, wang2017tacotron, ping2017deep}.
In particular, one of the most popular neural TTS pipeline~(a.k.a. ``end-to-end") consists of two components~\citep{ping2017deep, shen2018tacotron2}: (\emph{i}) an autoregressive \emph{seq2seq} model that generates mel~spectrogram from text, and (\emph{ii}) an autoregressive neural vocoder~(e.g., WaveNet) that synthesizes raw waveform  from mel~spectrogram.  This pipeline requires much less expert knowledge and only needs pairs of audio and transcript as training data.

However, the autoregressive nature of these models makes them quite slow at synthesis, because they operate sequentially at a high temporal resolution of waveform samples and spectrogram.
Most recently, several models are proposed for parallel waveform generation~\citep[e.g.,][]{oord2017parallel, ping2018clarinet, prenger2018waveglow, kumar2019melgan, binkowski2020high, ping2020waveflow}. 
In the end-to-end pipeline, the models (e.g., ClariNet, WaveFlow) still rely on autoregressive component to predict spectrogram features~(e.g., 100 frames per second).
In the linguistic feature-based pipeline, the models~(e.g., Parallel WaveNet, GAN-TTS) are conditioned on aligned linguistic features from phoneme duration model and F0 from frequency model, which are recurrent or autoregressive models.
Both of these TTS pipelines can be slow at synthesis on modern hardware optimized for parallel execution.


In this work, we present a fully parallel neural TTS system by proposing a non-autoregressive text-to-spectrogram model. 
Our major contributions are as follows:
\vspace{-0.3em}
\begin{enumerate}[itemsep=-0.00pt, topsep=0pt, leftmargin=1.5em]
    \item We propose ParaNet, a non-autoregressive attention-based architecture for text-to-speech, which is fully convolutional and converts text to mel spectrogram.
    It runs 254.6 times faster than real-time at synthesis on a 1080 Ti GPU, and brings $46.7$ times speed-up over its autoregressive counterpart~\citep{ping2017deep}, while obtaining reasonably good speech quality using neural vocoders.
    \item ParaNet distills the attention from the autoregressive text-to-spectrogram model, and iteratively refines the alignment between text and spectrogram in a layer-by-layer manner. It can produce more stable attentions than autoregressive Deep Voice~3~\citep{ping2017deep} on the challenging test sentences, because it does not have the discrepancy between the teacher-forced training and autoregressive inference.
    \item We build the fully parallel neural TTS system by combining ParaNet with parallel neural vocoder,  thus it can generate speech from text through a single feed-forward pass.
    We investigate several parallel vocoders, including the distilled IAF vocoder~\citep{ping2018clarinet} and WaveGlow~\citep{prenger2018waveglow}.
    To explore the possibility of training IAF vocoder~\emph{without distillation}, we also propose an alternative approach, WaveVAE, which can be trained from scratch within the variational autoencoder~(VAE) framework~\citep{kingma2013auto}.
\end{enumerate}
We organize the rest of paper as follows.
Section~\ref{sec:related_work} discusses related work.
We introduce the non-autoregressive ParaNet architecture in Section~\ref{sec:non-autoregressive}.
We discuss parallel neural vocoders  in Section~\ref{sec:wavevae}, and
 report experimental settings and results in Section~\ref{sec:experiment}.
We conclude the paper in Section~\ref{sec:conclusion}.

\section{Related work}
\label{sec:related_work}
%
Neural speech synthesis has obtained the state-of-the-art results and gained a lot of attention.
Several neural TTS systems were proposed, including WaveNet~\citep{oord2016wavenet}, Deep Voice~\citep{arik2017DV1}, Deep Voice~2~\citep{arik2017DV2}, Deep Voice~3~\citep{ping2017deep}, Tacotron~\citep{wang2017tacotron}, Tacotron 2~\citep{shen2018tacotron2}, Char2Wav~\citep{sotelo2017char2wav}, VoiceLoop~\citep{taigman2018voiceloop},
WaveRNN~\cite{kalchbrenner2018efficient}, ClariNet~\citep{ping2018clarinet}, and Transformer TTS~\citep{li2019neural}.
%
In particular, Deep~Voice~3, Tacotron and Char2Wav employ \emph{seq2seq} framework with the attention mechanism~\citep{bahdanau2014neural}, yielding much simpler pipeline compared to traditional multi-stage pipeline. 
Their excellent extensibility leads to promising results for several challenging tasks, such as voice cloning~\citep{arik2018neural, nachmani2018fitting, jia2018transfer, chen2018sample}.
All of these state-of-the-art systems are based on autoregressive models.

RNN-based autoregressive models, such as Tacotron and WaveRNN~\citep{kalchbrenner2018efficient}, lack parallelism at both training and synthesis. 
CNN-based autoregressive models, such as Deep~Voice~3 and WaveNet, enable parallel processing at training, but they still operate sequentially at synthesis since each output element must be generated before it can be passed in as input at the next time-step. 
%
Recently, there are some non-autoregressive models proposed for neural machine translation. 
\citet{gu2017non} trains a feed-forward neural network conditioned on fertility values, which are obtained from an external alignment system.
\citet{kaiser2018fast} proposes a latent variable model for fast decoding, while it remains autoregressiveness between latent variables. 
\citet{lee2018deterministic} iteratively refines the output sequence through a denoising autoencoder framework. 
Arguably, non-autoregressive model plays a more important role in text-to-speech, where the output speech spectrogram usually consists of hundreds of time-steps for a short text input with a few words. 
Our work is one of the first non-autoregressive \emph{seq2seq} model for TTS and provides as much as $46.7$ times speed-up at synthesis over its autoregressive counterpart~\citep{ping2017deep}.
There is a concurrent work~\citep{ren2019fastspeech}, which is based on the autoregressive transformer TTS~\citep{li2019neural} and can generate mel spectrogram in parallel. Our ParaNet is fully convolutional and lightweight. In contrast to FastSpeech, it has half of model parameters, requires smaller batch size~(16 vs. 64) for training and provides faster speed at synthesis~(see Table~\ref{tab:mos_tts} for detailed comparison).

Flow-based generative models~\citep{rezende2015variational, kingma2016improved, dinh2017density, kingma2018glow} transform a simple initial distribution into a more complex one by applying a series of invertible transformations. 
In previous work, flow-based models have obtained state-of-the-art results for parallel waveform synthesis~\citep{oord2017parallel, ping2018clarinet, prenger2018waveglow, kim2018flowavenet, yamamoto2019probability, ping2020waveflow}.

Variational autoencoder~(VAE)~\citep{kingma2013auto,rezende2014stochastic} has been applied for representation learning of natural speech for years. 
It models either the generative process of raw waveform~\citep{chung2015recurrent, van2017neural}, or spectrograms~\citep{hsu2018hierarchical}.
%
%
In previous work, autoregressive or recurrent neural networks are employed as the decoder of VAE~\citep{chung2015recurrent, van2017neural},  but they can be quite slow at synthesis. 
In this work, we employ a feed-forward IAF as the decoder, which enables parallel waveform synthesis.
%

%

\begin{figure*}[t!] \centering
\begin{tabular}{cc}
\hspace{-.1cm}
\includegraphics[height=3.4cm, clip]{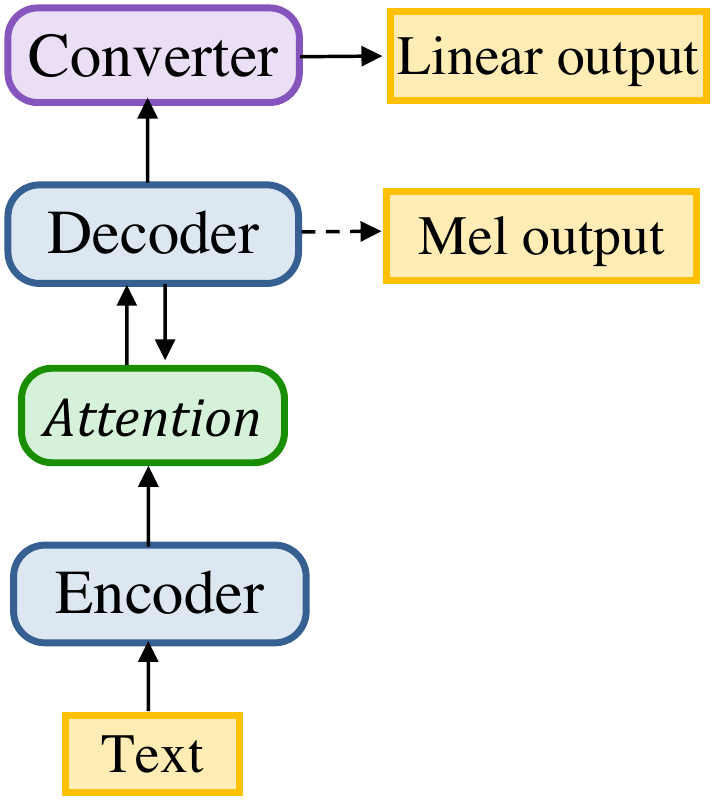} 
&\hspace{1.0cm}
\includegraphics[height=3.4cm, clip]{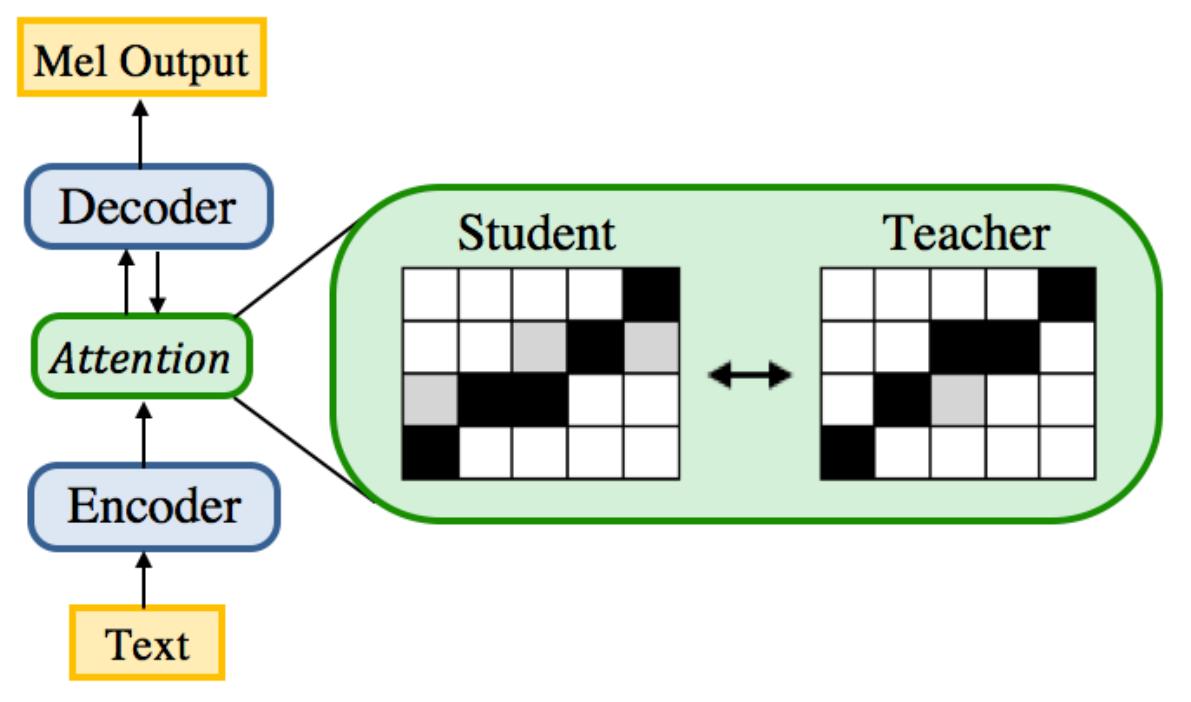} 
\\
\vspace{-.3em}
{\hspace{-.5em} \small (a)}  &  {\small(b)}
\\
\end{tabular}
\vspace{-.3em}
\caption{(a) Autoregressive \emph{seq2seq} model. The dashed line depicts the autoregressive decoding of mel spectrogram at inference.
(b) Non-autoregressive ParaNet model, which distills the attention from a pretrained autoregressive model.
}
\label{fig:encoder-decoder-concise} %
\end{figure*}

\begin{figure*}[t] \centering
\begin{tabular}{cc}
\hspace{-.3cm}
\includegraphics[height=4.3cm, clip]{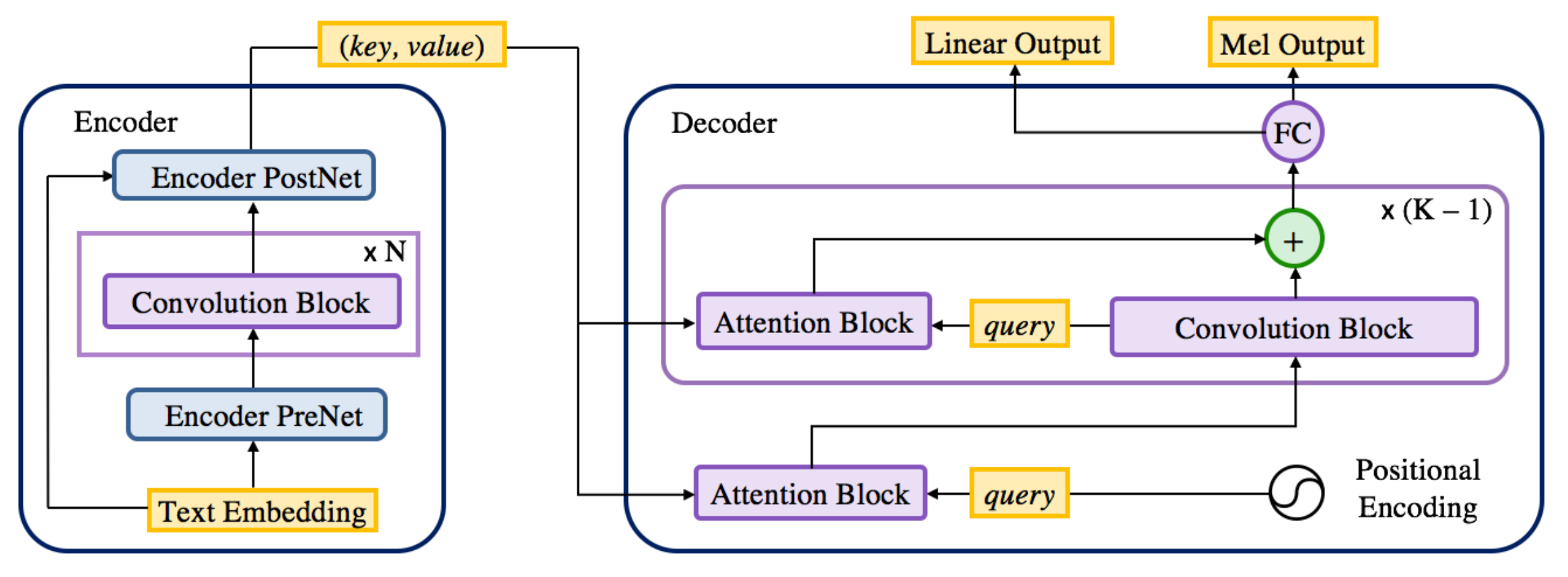}
&\hspace{1.2cm}
\includegraphics[height=4.5cm, clip]{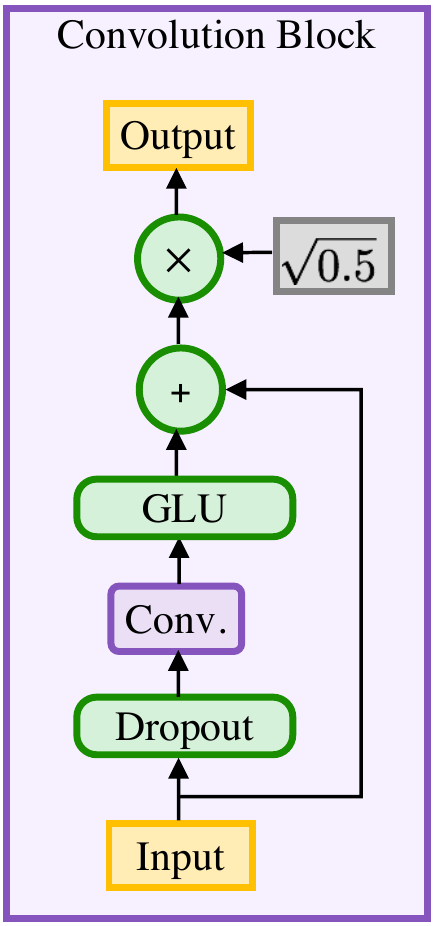} 
\\
\vspace{-.3em}
{\hspace{-.5em} \small (a)}  & \qquad\quad {\small(b)}
\\
\end{tabular}
\caption{(a)~Architecture of ParaNet. Its encoder provides \emph{key} and \emph{value} as the textual representation. The first attention block in decoder gets positional encoding as the \emph{query} and is followed by non-causal convolution blocks and attention blocks. (b) Convolution block appears in both encoder and decoder. It consists of a 1-D convolution with a gated linear unit~(GLU) and a residual
connection.}
\label{fig:encoder-decoder-detailed} %
\end{figure*}

\section{Text-to-spectrogram model}
\label{sec:non-autoregressive}
%
Our parallel TTS system has two components: 1) a feed-forward  text-to-spectrogram model, and 2) a parallel waveform synthesizer conditioned on mel spectrogram. 
In this section, we first present an autoregressive model derived from Deep Voice 3~(DV3)~\citep{ping2017deep}.
We then introduce ParaNet, a non-autoregressive text-to-spectrogram model~(see Figure~\ref{fig:encoder-decoder-concise}).

\subsection{Autoregressive architecture}
%

%
Our autoregressive model is based on DV3, a convolutional text-to-spectrogram architecture, which consists of three components:
\begin{itemize}[itemsep=0pt, topsep=0pt, leftmargin=1em]
    \item \textbf{Encoder}: A convolutional encoder, which takes text inputs and encodes them into internal hidden representation. 
    \item  \textbf{Decoder}: A \emph{causal} convolutional  decoder, which decodes the encoder representation with an \emph{attention} mechanism to log-mel spectragrams in an \emph{autoregressive} manner with an $\ell_1$ loss. 
    It starts with  a $1\times1$ convolution to preprocess the input log-mel spectrograms.
    \item  \textbf{Converter}: A \emph{non-causal} convolutional post processing network, which processes the hidden representation from the decoder using both past and future context information and predicts the log-linear spectrograms with an $\ell_1$ loss. 
    It enables bidirectional processing.
\end{itemize}
All these components use the same 1-D \emph{convolution block} with a gated linear unit as in DV3~(see Figure~\ref{fig:encoder-decoder-detailed}~(b) for more details).
The major difference between our model and DV3 is the decoder architecture.
The decoder of DV3 has multiple attention-based layers, where each layer consists of a causal convolution block followed by an attention block.
To simplify the \emph{attention distillation} described in Section~\ref{subsubsec:attention-distill}, our autoregressive decoder has only one attention block at its first layer. 
We find that reducing the number of attention blocks does not hurt the generated speech quality in general.

\subsection{Non-autoregressive architecture}
%
The proposed ParaNet~(see Figure~\ref{fig:encoder-decoder-detailed})  uses the same encoder architecture as the autoregressive model.
The decoder of ParaNet, conditioned solely on the hidden representation from the encoder, predicts the entire sequence of log-mel spectrograms in a feed-forward manner. 
As a result, both its training and synthesis can be done in parallel.
Specially, we make the following major architecture modifications from the autoregressive text-to-spectrogram model to the non-autoregressive model:
\begin{enumerate}[itemsep=0pt,topsep=0pt]
    \item \textbf{Non-autoregressive decoder}: Without the autoregressive generative constraint, the decoder can use \emph{non-causal} convolution blocks to take advantage of future context information and to improve model performance.
    In addition to log-mel spectrograms, it also predicts log-linear spectrograms with an $\ell_1$ loss for slightly better performance.
    We also remove the $1\times1$ convolution at the beginning, because the decoder does not take log-mel spectrograms as input.
    \item \textbf{No converter}: Non-autoregressive model removes the \emph{non-causal} converter since it already employs a \emph{non-causal} decoder. Note that, the major motivation of introducing non-causal \emph{converter} in  DV3 is to refine the decoder predictions based on bidirectional context information provided by non-causal convolutions. 
\end{enumerate}
\vspace{-0.3em}

\begin{figure*}[t] 
\centering
\hspace{-.3cm}
\includegraphics[height=4.0cm, clip]{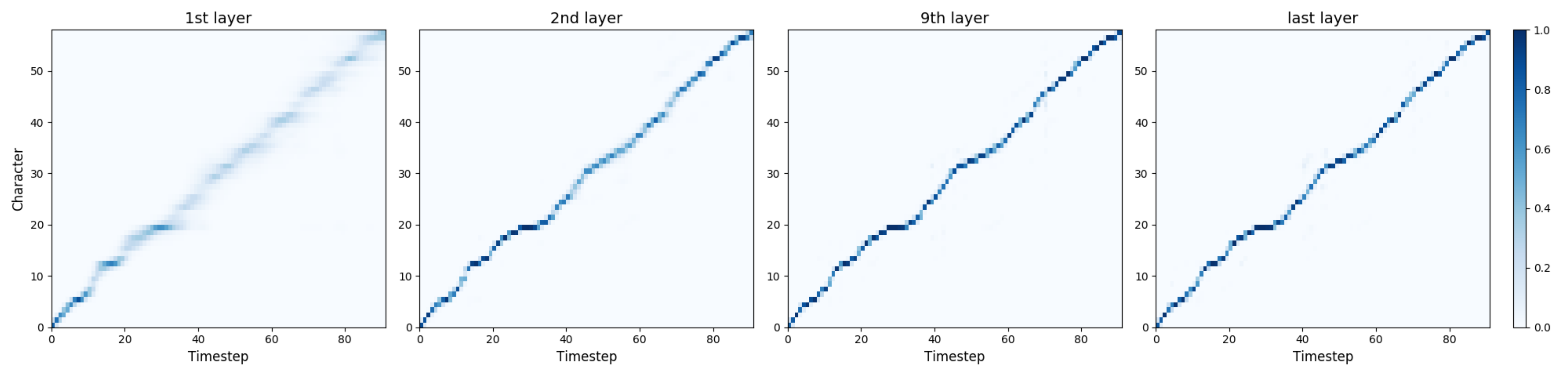}\\
\vspace{-0.2cm}
\caption{Our ParaNet iteratively refines the attention alignment in a  layer-by-layer way. One can see the 1st layer attention is mostly dominated by the positional encoding prior.
It becomes more and more confident about the alignment in the subsequent layers.}
\vspace{-0.1in}
\label{fig:attention_plot}
\end{figure*}

\subsection{Parallel attention mechanism}
It is challenging for the feed-forward model to learn the accurate alignment between the input text and output spectrogram. 
In particular, we need the full parallelism within the attention mechanism.
For example, the location-sensitive attention~\citep{chorowski2015attention, shen2018tacotron2} improves attention stability, but it performs sequentially at both training and synthesis, because it uses the cumulative attention weights from previous decoder time steps as an additional feature for the next time step.
Previous non-autoregressive decoders rely on an external alignment system~\citep{gu2017non}, or an autoregressive latent variable model~\citep{kaiser2018fast}.

In this work, we present several simple \& effective techniques, which could obtain accurate and stable attention alignment.
In particular, our non-autoregressive decoder can iteratively refine the attention alignment between text and mel spectrogram in a layer-by-layer manner as illustrated in Figure~\ref{fig:attention_plot}.
Specially, the decoder adopts a dot-product attention mechanism and consists of $K$ attention blocks (see Figure~\ref{fig:encoder-decoder-detailed}~(a)), where each attention block uses the per-time-step \emph{query} vectors from convolution block and per-time-step \emph{key} vectors from  encoder to compute the attention weights~\citep{ping2017deep}.
The attention block computes \emph{context} vectors as the weighted average of the \emph{value} vectors from the encoder. 
The non-autoregressive decoder starts with an attention block, in which the \emph{query} vectors are solely positional encoding~(see Section \ref{subsec:pos_encode} for details). 
The first attention block then provides the input for the convolution block at the next attention-based layer.

\subsubsection{Attention distillation}
\label{subsubsec:attention-distill}
%
We use the attention alignments from a pretrained autoregressive model to guide the training of non-autoregressive model.
Specifically, we minimize the cross entropy between the attention distributions from the non-autoregressive ParaNet and a pretrained autoregressive teacher.
We denote the attention weights from the non-autoregressive ParaNet as $W_{i, j}^{(k)}$, where $i$ and $j$ index the time-step of encoder and decoder respectively, and $k$ refers to the $k$-th attention block within the decoder. 
Note that, the attention weights $\{W_{i,j}^{(k)}\}_{i=1}^M$ form a valid distribution. 
We compute the \emph{attention~loss} as the average cross entropy  between the ParaNet and teacher's attention distributions:
\begin{equation}
l_{\text{atten}} = - \frac{1}{KN} \sum_{k=1}^{K} \sum_{j=1}^{N} \sum_{i=1}^{M}
W_{i,j}^{~t} \log W_{i,j}^{(k)},
\label{eq:attention_distill}
\end{equation}
\vspace{-0.1em}
\hspace{-0.3em}where $W_{i,j}^{~t}$ are the attention weights from the autoregressive teacher, $M$ and $N$ are the lengths of encoder and decoder, respectively. 
Our final loss function is a linear combination of $l_{\text{atten}}$ and $\ell_1$ losses from spectrogram predictions. 
We set the coefficient of $l_{\text{atten}}$ as $4$, and other coefficients as $1$ in all experiments.

\subsubsection{Positional encoding}
\label{subsec:pos_encode}
%
We use a similar positional encoding as in DV3 at every attention block~\citep{ping2017deep}. 
The positional encoding is added to both \emph{key} and \emph{query} vectors in the attention block, which forms an inductive bias for monotonic attention. 
Note that, the non-autoregressive model solely relies on its attention mechanism to decode mel spectrograms from the encoded textual features, without any autoregressive input. This makes the positional encoding even more crucial in guiding the attention to follow a monotonic progression over time at the beginning of training. 
The positional encodings  $h_p(i, k) =  \sin\left(\nicefrac{\omega_{s} i}{10000^{\nicefrac{k}{d}}}\right)$ (for even $i$), and $\cos\left(\nicefrac{\omega_{s} i}{10000^{\nicefrac{k}{d}}}\right)$ (for odd $i$), where $i$ is the time-step index, $k$ is the channel index, $d$ is the total number of channels in the positional encoding, and $\omega_s$ is the \emph{position rate} which indicates the average slope of the line in the attention distribution and roughly corresponds to the speed of speech.
We set $\omega_s$ in the following ways:
\begin{itemize}[itemsep=0pt,topsep=0pt,leftmargin=1.5em]
\item For the autoregressive teacher, $\omega_s$ is set to one for the positional encoding of \emph{query}. For the \emph{key}, it is set to the averaged ratio of the time-steps of spectrograms to the time-steps of textual features, which is around $6.3$ across our training dataset. Taking into account that a reduction factor of $4$ is used to simplify the learning of attention mechanism~\citep{wang2017tacotron} , $\omega_s$ is simply set as $\sfrac{6.3}{4}$ for the \emph{key} at both training and synthesis.
\vspace{-.25em}
\item For ParaNet, $\omega_s$ is also set to one for the \emph{query}, while $\omega_s$ for the \emph{key} is calculated differently. At training, $\omega_s$ is set to the ratio of the lengths of spectrograms and text for each individual training instance, which is also divided by a reduction factor of $4$. 
At synthesis, we need to specify the length of output spectrogram and the corresponding $\omega_s$, which actually controls the speech rate of the generated audios~(see Section II on demo website).
In all of our experiments, we simply set $\omega_s$ to be $\sfrac{6.3}{4}$ as in autoregressive model, and the length of output spectrogram as $\sfrac{6.3}{4}$ times the length of input text.
Such a setup yields an initial attention in the form of a diagonal line and guides the non-autoregressive decoder to refine its attention layer by layer~(see Figure~\ref{fig:attention_plot}).
\end{itemize}

\subsubsection{Attention masking}
Inspired by the attention masking in Deep Voice~3, we propose an attention masking scheme for the non-autoregressive ParaNet at synthesis:
\begin{itemize}[itemsep=0pt,topsep=0pt,leftmargin=1.5em]
\item For each \emph{query} from decoder, instead of computing the softmax over the entire set of encoder \emph{key} vectors, we compute the softmax only over a fixed window centered around the \emph{target position} and going forward and backward several time-steps~(e.g.,~3). The \emph{target position} is calculated as $\lfloor i_{\text{query}} \times 4 / 6.3 \rceil$, where $i_{\text{query}}$ is the time-step index of the \emph{query} vector, and $\lfloor\rceil$ is the rounding operator. 
\end{itemize}
We observe that this strategy reduces serious attention errors such as repeating or skipping words, and also yields clearer pronunciations, thanks to its more condensed attention distribution. Note that, this attention masking is shared across all attention blocks once it is generated, and does not prevent the parallel synthesis of the non-autoregressive model.

\section{Parallel waveform model}
\label{sec:wavevae}

As an indispensable component in our parallel neural TTS system, the parallel waveform model converts the mel spectrogram predicted from  ParaNet into the raw waveform. In this section, we discuss several existing parallel waveform models, and explore a new alternative in the system.

\subsection{Flow-based waveform models}
Inverse autoregressive flow (IAF)~\citep{kingma2016improved} is a special type of normalizing flow where each invertible transformation is based on an autoregressive neural network.
IAF performs synthesis in parallel and can easily reuse the expressive autoregressive architecture, such as WaveNet~\citep{oord2016wavenet}, which leads to the state-of-the-art results for speech synthesis~\citep{oord2017parallel, ping2018clarinet}.
However, the likelihood evaluation in IAF is autoregressive and slow, thus previous training methods rely on probability density distillation from a pretrained autoregressive WaveNet. 
This two-stage distillation process complicates the training pipeline and may  introduce pathological optimization~\citep{huangprobability}.

RealNVP~\citep{dinh2017density} and Glow~\citep{kingma2018glow} are different types of normalizing flows, where both synthesis and likelihood evaluation can be performed in parallel by enforcing bipartite architecture constraints.
Most recently, both of them were applied as parallel neural vocoders and can be trained from scratch~\citep{prenger2018waveglow, kim2018flowavenet}.
However, these models are less expressive than their autoregressive and IAF counterparts. One can find a detailed analysis in WaveFlow paper~\citep{ping2020waveflow}. 
In general, these bipartite flows require larger number of layers and hidden units, which lead to huge number of parameters. For example, a WaveGlow vocoder~\citep{prenger2018waveglow} has 87.88M parameters, whereas IAF vocoder has much smaller footprint with only 2.17M parameters~\citep{ping2018clarinet}, making it more preferred in production deployment.

\subsection{WaveVAE}
\label{subsec:wavevae}

Given the advantage of IAF vocoder, it is interesting to investigate whether it can be trained~\emph{without} the density distillation. 
One related work trains IAF within an auto-encoder~\citep{huangprobability}.
Our method uses the VAE framework, thus it is termed as WaveVAE. 
In contrast to~\citet{oord2017parallel} and \citet{ping2018clarinet}, WaveVAE can be trained from scratch by jointly optimizing the encoder $q_{\phi}(\vv z| \vv x, \vv c)$ and decoder $p_{\theta}(\vv x| \vv z, \vv c)$, where $\vv z$ is latent variables and $\vv c$ is the mel spectrogram conditioner.
We omit $\vv c$ for concise notation hereafter.

\subsubsection{Encoder}
The encoder of WaveVAE $q_{\phi} (\vv z | \vv x)$ is parameterized by a Gaussian autoregressive WaveNet~\citep{ping2018clarinet} that maps the ground truth audio $\vv x$ into the same length latent representation $\vv z$.
Specifically, the Gaussian WaveNet models $x_t$ given the previous samples $x_{<t}$ as 
$x_t \sim \mathcal{N} \big( \mu (x_{<t}; \phi), \sigma (x_{<t}; \phi) \big), $
where the mean $\mu (x_{<t}; \phi)$ and scale $\sigma (x_{<t}; \phi)$ are predicted by WaveNet, respectively.
The encoder posterior is constructed as,
\begin{align*}
&q_{\phi} (\vv z | \vv x) = \prod_{t} q_{\phi} ( z_t ~|~  x_{\le t} ), \\
&~\text{where}~\ 
q_{\phi} ( z_t ~|~  x_{\le t}) = \mathcal{N} 
\big( \frac{x_t - \mu(x_{<t}; \phi)}{\sigma(x_{<t}; \phi)}, \varepsilon \big).
\label{eq:encoder}
\end{align*}
Note that, the mean $\mu(x_{<t}; \phi)$ and scale $\sigma(x_{<t})$ are applied for ``whitening'' the posterior distribution. 
We introduce a trainable scalar $\varepsilon>0$ to decouple the  global variation, which will make optimization process easier. 
Given the observed $\vv x$, the $q_{\phi} (\vv z | \vv x)$ admits parallel sampling of latents $\vv z$. 
One can build the connection between the encoder of WaveVAE and the teacher model of ClariNet, as both of them use a Gaussian WaveNet to guide the training of IAF for parallel wave generation.

\subsubsection{Decoder}
Our decoder $p_{\theta} (\vv x | \vv z )$ is parameterized by the one-step-ahead predictions from an IAF~\citep{ping2018clarinet}.
We let $\vv z^{(0)} = \vv z$ and apply a stack of IAF transformations from $\vv z^{(0)} \rightarrow  \dots \vv z^{(i)} \rightarrow \dots \vv z^{(n)}$, and each transformation $\vv z^{(i)} = f(\vv z^{(i-1)}; \theta)$  is defined as,
\begin{align}
\vv z^{(i)} = \vv z^{(i-1)} \cdot \vv \sigma^{(i)}
+ \vv \mu^{(i)},
\end{align}
where $\mu_t^{(i)} = \mu(z_{<t}^{(i-1)}; \theta)$ and $\sigma_t^{(i)} = \sigma(z_{<t}^{(i-1)}; \theta)$  are shifting and scaling variables  modeled by a Gaussian WaveNet.
One can show that, given $\vv z^{(0)} \sim \mathcal{N}(\vv\mu^{(0)}, \vv\sigma^{(0)})$  from the Gaussian prior or encoder, the per-step $p(z_t^{(n)} \mid z_{<t}^{(0)})$ also follows Gaussian with scale and mean as,
\begin{align}
 {\vv \sigma}^{\text{tot}} =  \prod_{i=0}^{n} \vv \sigma^{(i)},\quad
 {\vv \mu}^{\text{tot}} = \sum_{i=0}^{n} \vv \mu^{(i)} \prod_{j>i}^{n} \vv \sigma^{(j)}.
\end{align}
Lastly, we set
$\vv x = \vv \epsilon \cdot  {\vv \sigma}^{\text{tot}} 
+ {\vv \mu}^{\text{tot}}$, 
where $\vv \epsilon \sim \mathcal{N}(0, I)$.
Thus, $p_{\theta}(\vv x ~|~ \vv z) = \mathcal{N}( {\vv \mu}^{\text{tot}},  {\vv \sigma}^{\text{tot}} )$.
For the generative process, we use the standard Gaussian prior $p(\vv z) = \mathcal{N}(0, I)$.  

\subsubsection{Training objective}
We maximize the evidence lower bound (ELBO) for observed $\vv x$ in VAE,
\begin{equation}
\max_{\phi, \theta} \
\mathbb{E}_{q_{\phi} (\vv z | \vv x )} 
\big[\log p_{\theta} (\vv x | \vv z ) \big] -
 \text{KL} \big( q_{\phi} (\vv z | \vv x ) \,||\, p(\vv z) \big),
\label{eq:obj_vae}
\end{equation}
where the KL divergence can be calculated in closed-form as both $q_{\phi} (\vv z | \vv x )$ and $p(\vv z)$ are Gaussians,
\begin{align*}
&\text{KL} \big( q_{\phi} (\vv z | \vv x) \,||\, p(\vv z) \big) \\
= &\sum_{t} \log\frac{1}{\varepsilon} + \frac{1}{2} \Big( \varepsilon^2 - 1 + 
\big( \frac{x_t - \mu(x_{<t})}{\sigma(x_{<t})} \big)^2  \Big).
\label{eq:kld}
\end{align*}
The reconstruction term in Eq.~\eqref{eq:obj_vae} is  intractable to compute exactly.
We do stochastic optimization by drawing a sample $\vv z$ from the encoder $q_{\phi}(\vv z|\vv x)$ through the reparameterization trick, and evaluating the likelihood $\log p_{\theta} (\vv x | \vv z )$.
To avoid the ``posterior collapse'', in which the posterior distribution $q_{\phi} (\vv z | \vv x )$ quickly collapses to the white noise prior $p(\vv z)$ at the early stage of training, we apply the annealing strategy for KL divergence, where its weight is gradually increased from 0 to 1, via a sigmoid function~\citep{bowman2016generating}.
Through it, the encoder can encode sufficient information into the latent representations at the early training, and then gradually regularize the latent representation by increasing the weight of the KL divergence.

{\bf STFT loss:}
Similar to~\citet{ping2018clarinet}, we also add a short-term Fourier transform (STFT) loss to improve the quality of synthesized speech.
We define the \emph{STFT loss} as the summation of $\ell_2$ loss on the magnitudes of STFT and $\ell_1$ loss on the log-magnitudes of STFT between the output audio and ground truth audio~\citep{ping2018clarinet, arik2019fast, wang2019neural}.
For STFT, we use a $12.5$ms frame-shift, $50$ms Hanning window length, and we set the FFT size to $2048$. 
We consider two STFT losses in our objective: $(i)$~the STFT loss between ground truth audio and reconstructed audio using encoder $q_{\phi}(\vv z| \vv x)$; 
$(ii)$ the STFT loss between ground truth audio and synthesized audio using the prior $p(\vv z)$, with the purpose of reducing the gap between reconstruction and synthesis. 
Our final loss is a linear combination of VAE objective in Eq.~\eqref{eq:obj_vae} and the STFT losses. The corresponding coefficients are simply set to be one in all of our experiments.

\section{Experiment}
\label{sec:experiment}
In this section, we present several experiments to evaluate the proposed ParaNet and WaveVAE.

\begin{table*}[h]
\caption{Hyperparameters of autoregressive text-to-spectrogram model and non-autoregressive ParaNet in the experiment.}
\vspace{0.3em}
\centering
\begin{tabular}{|c|c|c|}
\hline
\textbf{Hyperparameter}                     &   \textbf{Autoregressive Model} & \textbf{Non-autoregressive Model}  \\ \hline
FFT Size                               &      2048          &  2048     \\  \hline
FFT Window Size / Shift                &      1200 / 300    &  1200 / 300    \\ \hline
Audio Sample Rate                      &        24000      & 24000 \\ \hline
Reduction Factor $r$                   &         4         & 4           \\ \hline
Mel Bands                              &         80        & 80                 \\ \hline
Character Embedding Dim.                &        256       & 256  \\ \hline
Encoder Layers / Conv. Width / Channels &      7 / 5 / 64  &  7 / 9 / 64         \\ \hline
Decoder PreNet Affine Size                  &      128, 256      & N/A               \\ \hline
Decoder Layers / Conv. Width            &        4 / 5  &     17 / 7             \\ \hline
Attention Hidden Size                  &        128     & 128   \\ \hline
Position Weight / Initial Rate         &        1.0 / 6.3     & 1.0 / 6.3         \\ \hline
PostNet Layers / Conv. Width / Channels &      5 / 5 / 256  & N/A           \\ \hline
Dropout Keep Probability                    &        0.95    &   1.0     \\   \hline
ADAM Learning Rate                     &            0.001   &   0.001                  \\ \hline
Batch Size                             &            16      &   16            \\ \hline
Max Gradient Norm                      &            100     &   100                \\ \hline
Gradient Clipping Max. Value            &           5.0     &   5.0                   \\ \hline
Total Number of Parameters            &           6.85M     &   17.61M                   \\ \hline
\end{tabular}
\label{tab:hyperparameters}
\end{table*}

\begin{table*}[t]
\centering
\caption{The model footprint, synthesis time for 1 second speech~(on 1080Ti with FP32), and the 5-scale Mean Opinion Score (MOS) ratings with 95\% confidence intervals for comparison.}
\vspace{0.3em}
\begin{tabular}{l|r|r|r}
\hline
\textbf{Neural TTS system} & \textbf{\# parameters}  & \textbf{synthesis time~(ms)} & \textbf{MOS score} \\ \hline
DV3 + WaveNet &  6.85 + 9.08 = 15.93 M &  181.8 ~+~ 5$\times10^5$ &  \quad  ${4.09 \pm 0.26}$  \\
ParaNet + WaveNet &  17.61 + 9.08 = 26.69 M & 3.9 ~+~ 5$\times10^5$ &  \quad $4.01 \pm 0.24$  \\ 
DV3 + ClariNet & 6.85 + 2.17  =\ \ \ 9.02~M & 181.8 ~+~ 64.9 &  \quad $3.88 \pm 0.25$  \\
ParaNet +  ClariNet  & 17.61 + 2.17 = {\bf 19.78}~M & {\bf 3.9 ~+~ 64.9} & \quad {\bf $3.62 \pm 0.23$} \\
DV3 + WaveVAE &   6.85 + 2.17  =\ \ \ 9.02~M & 181.8 ~+~ 64.9 & \quad $3.70 \pm 0.29$  \\
ParaNet + WaveVAE & 17.61 + 2.17 = 19.78~M & 3.9 ~+~ 64.9 & \quad $3.31 \pm 0.32$ \\
DV3 + WaveGlow &   6.85 + 87.88 =  94.73~M &  181.8 ~+~ 117.6 & \quad $3.92 \pm 0.24$  \\
ParaNet + WaveGlow & 17.61 + 87.88 = 105.49~M & 3.9 ~+~ 117.6 & \quad $3.27 \pm 0.28$ \\
FastSpeech~(re-impl) + WaveGlow & 31.77 + 87.88 = {\bf 119.65}~M & {\bf 6.2 ~+~ 117.6} & \quad \emph{3.56 $\pm$ 0.26} \\
\hline
\end{tabular}
\label{tab:mos_tts}
\vspace{-.6em}
\end{table*}


\begin{table*}[t]
\centering
\caption{Attention error counts for text-to-spectrogram models on the 100-sentence test set.
One or more mispronunciations, skips, and repeats count as a single mistake per utterance. 
The non-autoregressive ParaNet~(17-layer decoder) with attention mask obtains the fewest attention errors in total. For ablation study, we include the results for two additional  ParaNet models. They have 6 and 12 decoder layers and are denoted as ParaNet-6 and ParaNet-12, respectively.}
\vspace{0.3em}
\begin{tabular}{lc|ccc|c}
\hline
\textbf{Model} & \hspace{-0.4em}\textbf{Attention mask}          & \textbf{Repeat} & \hspace{-0.7em} \textbf{Mispronounce} & \hspace{-0.7em} \textbf{Skip}  & \hspace{-0.7em} \textbf{Total} \\ \hline
\hspace{-0.3em} Deep Voice 3  & \multirow{4}{*}{No}  & 12      & 10     & 15   &  37  \\
\hspace{-0.3em} {ParaNet} &   & \textbf{1} & \textbf{4}  & \textbf{7} & \textbf{12}  \\
\hspace{-0.3em} ParaNet-12 &  & 5 & 7  & 5 & 17  \\
\hspace{-0.3em} ParaNet-6 & & 4 & 11  & 11 & 26  \\
\hline
\hspace{-0.3em} Deep Voice 3 & \multirow{4}{*}{Yes} & 1 & 4 & 3 &  8 \\
\hspace{-0.3em} {ParaNet} &  & \textbf{2} & \textbf{4} & \textbf{0} & \textbf{6} \\
\hspace{-0.3em} ParaNet-12 &  & 4 & 6 & 2 & 12 \\
\hspace{-0.3em} ParaNet-6 & & 3 & 10 & 3 & 16 \\ \hline
\end{tabular}
\label{tab:attention_errors}
\vspace{-.3em}
\end{table*}

\subsection{Settings}
{\bf Data}: In our experiment, we use an internal English speech dataset containing about 20 hours of speech data from a female speaker with a sampling rate of 48~kHz.
We downsample the audios to 24~kHz.

{\bf Text-to-spectrogram models:}
For both ParaNet and Deep Voice 3~(DV3), we use the mixed representation of characters and phonemes~\citep{ping2017deep}.
The default hyperparameters of ParaNet and DV3 are provided in Table~\ref{tab:hyperparameters}.
Both ParaNet and DV3 are trained for 500K steps using Adam optimizer~\citep{kingma2014adam}.
We find that larger kernel width and deeper layers generally help improve the performance of ParaNet.
In terms of the number of parameters,  our ParaNet~({17.61~M} params) is  ${2.57}\times$  larger than the Deep Voice 3~({6.85M} params) and  $1.71\times$ smaller than the FastSpeech~(30.1M params)~\citep{ren2019fastspeech}.
We use an open source reimplementation of FastSpeech~\footnote{\url{https://github.com/xcmyz/FastSpeech}} by adapting the hyperparameters for handling the 24kHz dataset.

{\bf Neural vocoders:}
In this work, we compare various neural vocoders paired with text-to-spectrogram models, including WaveNet~\citep{oord2016wavenet}, ClariNet~\citep{ping2018clarinet}, WaveVAE, and WaveGlow~\citep{prenger2018waveglow}.
We train all neural vocoders on 8 Nvidia 1080Ti GPUs using randomly chosen 0.5s audio clips.

We train two 20-layer WaveNets with residual channel 256 conditioned on the predicted mel spectrogram from ParaNet and DV3, respectively.
We apply two layers of convolution block to process the predicted mel spectrogram, and use two layers of transposed 2-D convolution (in time and frequency) interleaved with
leaky ReLU ($\alpha = 0.4$) to upsample the outputs from frame-level to sample-level.
We use the Adam optimizer~\citep{kingma2014adam} with a batch size of 8 and a learning rate of 0.001 at the beginning, which is annealed by half every 200K steps.
We train the models for 1M steps.

We use the same IAF architecture as ClariNet~\citep{ping2018clarinet}.
It consists of four stacked Gaussian IAF blocks, which are parameterized by [10, 10, 10, 30]-layer WaveNets respectively, with the 64 residual \& skip channels and filter size 3 in dilated convolutions.
The IAF is conditioned on log-mel spectrograms with two layers of transposed 2-D convolution as in ClariNet.
We use the same teacher-student setup for ClariNet as in \citet{ping2018clarinet} and we train a 20-layer Gaussian autoregressive WaveNet as the teacher model.
For the encoder in WaveVAE, we also use a 20-layers Gaussian WaveNet conditioned on  log-mel spectrograms.
For the decoder, we use the same architecture as the distilled IAF.
Both the encoder and decoder of WaveVAE share the same conditioner network.
Both of the distilled IAF and WaveVAE are trained on ground-truth mel spectrogram. 
We use Adam optimizer with 1000K steps for distilled IAF. For WaveVAE, we train it for 400K because it converges much faster.
The learning rate is set to 0.001 at the beginning and annealed by half every 200K steps for both models.

We use the open source implementation of WaveGlow with default hyperparameters~(residual channel 256)~\footnote{\url{https://github.com/NVIDIA/waveglow}}, except change the sampling rate from 22.05kHz to 24kHz, FFT window length from 1024 to 1200, and FFT window shift from 256 to 300 for handling the 24kHz dataset. The model is trained for 2M steps.

\subsection{Results}

{\bf{Speech quality:}}
We use the crowdMOS toolkit~\citep{ribeiro2011crowdmos} for subjective Mean Opinion Score (MOS) evaluation.
We report the MOS results in Table~\ref{tab:mos_tts}.
The ParaNet can provide comparable quality of speech as the autoregressive DV3 using WaveNet vocoder~(MOS: 4.09 vs. 4.01).
When we use the ClariNet vocoder, ParaNet can still provide reasonably good speech quality~(MOS: 3.62) as a fully feed-forward TTS system. 
WaveVAE obtains worse results than distilled IAF vocoder, but it can be trained from scratch and simplifies the training pipeline.
When conditioned on predicted mel spectrogram, WaveGlow tends to produce constant frequency artifacts. 
To remedy this, we applied the denoising function with strength 0.1, as recommended in the repository of WaveGlow.
It is effective when the predicted mel spectrograms are from DV3, but not effective when the predicted mel spectrograms are from ParaNet. As a result, the MOS score degrades seriously.
We add the comparison with FastSpeech after the paper submission. 
Because it is costly to relaunch the MOS evaluations of all the models, we perform a separate MOS evaluation for FastSpeech. 
Note that, the group of human raters can be different on Mechanical Turk, and the subjective scores may not be directly comparable.
{One can find the synthesized speech samples in: \url{https://parallel-neural-tts-demo.github.io/} .}

{\bf{Synthesis speed:}}
We test synthesis speed of all models on NVIDIA GeForce GTX 1080~Ti with 32-bit floating point (FP32) arithmetic.
We compare the ParaNet with the autoregressive DV3 in terms of inference latency. We construct a custom 15-sentence test set (see Appendix~A) and run inference for 50 runs on each of the 15 sentences (batch size is set to $1$). The average audio duration of the utterances is 6.11 seconds.
The average inference latencies over 50 runs and 15 sentences are $0.024$ and $1.12$ seconds for ParaNet and DV3, respectively. Hence, our ParaNet runs 254.6 times faster than real-time and brings about $46.7$ times speed-up over its small-footprint autoregressive counterpart at synthesis.
It also runs 1.58 times faster than FastSpeech.
We summarize synthesis speed of TTS systems in Table~\ref{tab:mos_tts}.
One can observe that the latency bottleneck is the autoregressive text-to-spectrogram model, when the system uses parallel neural vocoder.
The ClariNet and WaveVAE vocoders have much smaller footprint and faster synthesis speed than WaveGlow. 

{\bf Attention error analysis: }
In autoregressive models, there is a noticeable discrepancy between the teacher-forced training and autoregressive inference, which can yield accumulated errors along the generated sequence at synthesis~\citep{bengio2015scheduled}.
In neural TTS, this discrepancy leads to miserable attention errors at autoregressive inference, including (\emph{i}) repeated words, (\emph{ii}) mispronunciations, and (\emph{iii}) skipped words~(see \citet{ping2017deep} for detailed examples), which is a critical problem for online deployment of attention-based neural TTS systems. 
We perform an attention error analysis for our non-autoregressive ParaNet on a 100-sentence test set~(see Appendix~B), which includes particularly-challenging cases from deployed TTS systems (e.g. dates, acronyms, URLs, repeated words, proper nouns, and foreign words).
In Table~\ref{tab:attention_errors}, we find that the non-autoregressive ParaNet has much fewer attention errors than its autoregressive counterpart at synthesis~($12$ vs. $37$) without attention mask.
Although our ParaNet distills the (teacher-forced) attentions from an autoregressive model, it only takes textual inputs at both training and synthesis and does not have the similar discrepancy as in autoregressive model.
%
%
In previous work, attention masking was applied to enforce the monotonic attentions and reduce attention errors, and was demonstrated to be effective in Deep~Voice~3~\citep{ping2017deep}.
We find that our non-autoregressive ParaNet still can have  fewer attention errors than autoregressive DV3~($6$ vs. $8$), when both of them use the attention masking.

\subsection{Ablation study}
We perform ablation studies to verify the effectiveness of several techniques used in ParaNet, including attention distillation, positional encoding, and stacking decoder layers to refine the attention alignment in a layer-by-layer manner.
We evaluate the performance of a non-autoregressive ParaNet model trained without attention distillation and find that it fails to learn meaningful attention alignment. The synthesized audios are unintelligible and mostly pure noise.
Similarly, we train another non-autoregressive ParaNet model without adding positional encoding in the attention block. The resulting model only learns very blurry attention alignment and cannot synthesize intelligible speech.
Finally, we train two non-autoregressive ParaNet models with 6 and 12 decoder layers, respectively, and compare them with the default non-autoregressive ParaNet model which has 17 decoder layers. We conduct the same attention error analysis on the 100-sentence test set and the results are shown in Table~\ref{tab:attention_errors}. We find that increasing the number of decoder layers for non-autoregressive ParaNet can reduce the total number of attention errors, in both cases with and without applying attention mask at synthesis.

\section{Conclusion}
\label{sec:conclusion}
In this work, we build a feed-forward neural TTS system by proposing a non-autoregressive text-to-spectrogram model.
The proposed ParaNet obtains reasonably good speech quality and brings $46.7$ times speed-up over its autoregressive counterpart at synthesis.
We also compare various neural vocoders within the TTS system.
Our results suggest that the parallel vocoder is generally less robust than WaveNet vocoder, when the front-end acoustic model is non-autoregressive.
As a result, it is interesting to investigate small-footprint and robust parallel neural vocoder~(e.g., WaveFlow) in future study.

\bibliography{reference}
\bibliographystyle{icml2020}

\onecolumn

\appendix

\section*{{\Large Appendix}}

\section{15-Sentence Test Set}
\label{sec:15-test}
The 15 sentences used to quantify the inference speed up in Table~\ref{tab:mos_tts} are listed below (note that \% corresponds to pause):

1. WHEN THE SUNLIGHT STRIKES RAINDROPS IN THE AIR\%THEY ACT AS A PRISM AND FORM A RAINBOW\%.\\
2. THESE TAKE THE SHAPE OF A LONG ROUND ARCH\%WITH ITS PATH HIGH ABOVE\%AND ITS TWO ENDS APPARENTLY BEYOND THE HORIZON\%.\\
3. WHEN A MAN LOOKS FOR SOMETHING BEYOND HIS REACH\%HIS FRIENDS SAY HE IS LOOKING FOR THE POT OF GOLD AT THE END OF THE RAINBOW\%.\\
4. IF THE RED OF THE SECOND BOW FALLS UPON THE GREEN OF THE FIRST\%THE RESULT IS TO GIVE A BOW WITH AN ABNORMALLY WIDE YELLOW BAND\%.\\
5. THE ACTUAL PRIMARY RAINBOW OBSERVED IS SAID TO BE THE EFFECT OF SUPER IMPOSITION OF A NUMBER OF BOWS\%.\\
6. THE DIFFERENCE IN THE RAINBOW DEPENDS CONSIDERABLY UPON THE SIZE OF THE DROPS\%.\\
7. IN THIS PERSPECTIVE\%WE HAVE REVIEWED SOME OF THE MANY WAYS IN WHICH NEUROSCIENCE HAS MADE FUNDAMENTAL CONTRIBUTIONS\%.\\
8. IN ENHANCING AGENT CAPABILITIES\%IT WILL BE IMPORTANT TO CONSIDER OTHER SALIENT PROPERTIES OF THIS PROCESS IN HUMANS\%.\\
9. IN A WAY THAT COULD SUPPORT DISCOVERY OF SUBGOALS AND HIERARCHICAL PLANNING\%.\\
10. DISTILLING INTELLIGENCE INTO AN ALGORITHMIC CONSTRUCT AND COMPARING IT TO THE HUMAN BRAIN MIGHT YIELD INSIGHTS\%.\\
11. THE VAULT THAT WAS SEARCHED HAD IN FACT BEEN EMPTIED EARLIER THAT SAME DAY\%.\\
12. ANT LIVES NEXT TO GRASSHOPPER\%ANT SAYS\%I LIKE TO WORK EVERY DAY\%.\\
13. YOUR MEANS OF TRANSPORT FULFIL ECONOMIC REQUIREMENTS IN YOUR CHOSEN COUNTRY\%.\\
14. SLEEP STILL FOGGED MY MIND AND ATTEMPTED TO FIGHT BACK THE PANIC\%.\\
15. SUDDENLY\%I SAW TWO FAST AND FURIOUS FEET DRIBBLING THE BALL TOWARDS MY GOAL\%.\\

\section{100-Sentence Test Set}
\label{sec:100-test}
The 100 sentences used to quantify the results in Table~\ref{tab:attention_errors} are listed below (note that \% corresponds to pause):

1. A B C\%.\\
2. X Y Z\%.\\
3. HURRY\%.\\
4. WAREHOUSE\%.\\
5. REFERENDUM\%.\\
6. IS IT FREE\%?\\
7. JUSTIFIABLE\%.\\
8. ENVIRONMENT\%.\\
9. A DEBT RUNS\%.\\
10. GRAVITATIONAL\%.\\
11. CARDBOARD FILM\%.\\
12. PERSON THINKING\%.\\
13. PREPARED KILLER\%.\\
14. AIRCRAFT TORTURE\%.\\
15. ALLERGIC TROUSER\%.\\
16. STRATEGIC CONDUCT\%.\\
17. WORRYING LITERATURE\%.\\
18. CHRISTMAS IS COMING\%.\\
19. A PET DILEMMA THINKS\%.\\
20. HOW WAS THE MATH TEST\%?\\
21. GOOD TO THE LAST DROP\%.\\
22. AN M B A AGENT LISTENS\%.\\
23. A COMPROMISE DISAPPEARS\%.\\
24. AN AXIS OF X Y OR Z FREEZES\%.\\
25. SHE DID HER BEST TO HELP HIM\%.\\
26. A BACKBONE CONTESTS THE CHAOS\%.\\
27. TWO A GREATER THAN TWO N NINE\%.\\
28. DON'T STEP ON THE BROKEN GLASS\%.\\
29. A DAMNED FLIPS INTO THE PATIENT\%.\\
30. A TRADE PURGES WITHIN THE B B C\%.\\
31. I'D RATHER BE A BIRD THAN A FISH\%.\\
32. I HEAR THAT NANCY IS VERY PRETTY\%.\\
33. I WANT MORE DETAILED INFORMATION\%.\\
34. PLEASE WAIT OUTSIDE OF THE HOUSE\%.\\
35. N A S A EXPOSURE TUNES THE WAFFLE\%.\\
36. A MIST DICTATES WITHIN THE MONSTER\%.\\
37. A SKETCH ROPES THE MIDDLE CEREMONY\%.\\
38. EVERY FAREWELL EXPLODES THE CAREER\%.\\
39. SHE FOLDED HER HANDKERCHIEF NEATLY\%.\\
40. AGAINST THE STEAM CHOOSES THE STUDIO\%.\\
41. ROCK MUSIC APPROACHES AT HIGH VELOCITY\%.\\
42. NINE ADAM BAYE STUDY ON THE TWO PIECES\%.\\
43. AN UNFRIENDLY DECAY CONVEYS THE OUTCOME\%.\\
44. ABSTRACTION IS OFTEN ONE FLOOR ABOVE YOU\%.\\
45. A PLAYED LADY RANKS ANY PUBLICIZED PREVIEW\%.\\
46. HE TOLD US A VERY EXCITING ADVENTURE STORY\%.\\
47. ON AUGUST TWENTY EIGTH\%MARY PLAYS THE PIANO\%.\\
48. INTO A CONTROLLER BEAMS A CONCRETE TERRORIST\%.\\
49. I OFTEN SEE THE TIME ELEVEN ELEVEN ON CLOCKS\%.\\
50. IT WAS GETTING DARK\%AND WE WEREN'T THERE YET\%.\\
51. AGAINST EVERY RHYME STARVES A CHORAL APPARATUS\%.\\
52. EVERYONE WAS BUSY\%SO I WENT TO THE MOVIE ALONE\%.\\
53. I CHECKED TO MAKE SURE THAT HE WAS STILL ALIVE\%.\\
54. A DOMINANT VEGETARIAN SHIES AWAY FROM THE G O P\%.\\
55. JOE MADE THE SUGAR COOKIES\%SUSAN DECORATED THEM\%.\\
56. I WANT TO BUY A ONESIE\%BUT KNOW IT WON'T SUIT ME\%.\\
57. A FORMER OVERRIDE OF Q W E R T Y OUTSIDE THE POPE\%.\\
58. F B I SAYS THAT C I A SAYS\%I'LL STAY AWAY FROM IT\%.\\
59. ANY CLIMBING DISH LISTENS TO A CUMBERSOME FORMULA\%.\\
60. SHE WROTE HIM A LONG LETTER\%BUT HE DIDN'T READ IT\%.\\
61. DEAR\%BEAUTY IS IN THE HEAT NOT PHYSICAL\%I LOVE YOU\%.\\
62. AN APPEAL ON JANUARY FIFTH DUPLICATES A SHARP QUEEN\%.\\
63. A FAREWELL SOLOS ON MARCH TWENTY THIRD SHAKES NORTH\%.\\
64. HE RAN OUT OF MONEY\%SO HE HAD TO STOP PLAYING POKER\%.\\
65. FOR EXAMPLE\%A NEWSPAPER HAS ONLY REGIONAL DISTRIBUTION T\%.\\
66. I CURRENTLY HAVE FOUR WINDOWS OPEN UP\%AND I DON'T KNOW WHY\%.\\
67. NEXT TO MY INDIRECT VOCAL DECLINES EVERY UNBEARABLE ACADEMIC\%.\\
68. OPPOSITE HER SOUNDING BAG IS A M C'S CONFIGURED THOROUGHFARE\%.\\
69. FROM APRIL EIGHTH TO THE PRESENT\%I ONLY SMOKE FOUR CIGARETTES\%.\\
70. I WILL NEVER BE THIS YOUNG AGAIN\%EVER\%OH DAMN\%I JUST GOT OLDER\%.\\
71. A GENEROUS CONTINUUM OF AMAZON DOT COM IS THE CONFLICTING WORKER\%.\\
72. SHE ADVISED HIM TO COME BACK AT ONCE\%THE WIFE LECTURES THE BLAST\%.\\
73. A SONG CAN MAKE OR RUIN A PERSON'S DAY IF THEY LET IT GET TO THEM\%.\\
74. SHE DID NOT CHEAT ON THE TEST\%FOR IT WAS NOT THE RIGHT THING TO DO\%.\\
75. HE SAID HE WAS NOT THERE YESTERDAY\%HOWEVER\%MANY PEOPLE SAW HIM THERE\%.\\
76. SHOULD WE START CLASS NOW\%OR SHOULD WE WAIT FOR EVERYONE TO GET HERE\%?\\
77. IF PURPLE PEOPLE EATERS ARE REAL\%WHERE DO THEY FIND PURPLE PEOPLE TO EAT\%?\\
78. ON NOVEMBER EIGHTEENTH EIGHTEEN TWENTY ONE\%A GLITTERING GEM IS NOT ENOUGH\%.\\
79. A ROCKET FROM SPACE X INTERACTS WITH THE INDIVIDUAL BENEATH THE SOFT FLAW\%.\\
80. MALLS ARE GREAT PLACES TO SHOP\%I CAN FIND EVERYTHING I NEED UNDER ONE ROOF\%.\\
81. I THINK I WILL BUY THE RED CAR\%OR I WILL LEASE THE BLUE ONE\%THE FAITH NESTS\%.\\
82. ITALY IS MY FAVORITE COUNTRY\%IN FACT\%I PLAN TO SPEND TWO WEEKS THERE NEXT YEAR\%.\\
83. I WOULD HAVE GOTTEN W W W DOT GOOGLE DOT COM\%BUT MY ATTENDANCE WASN'T GOOD ENOUGH\%.\\
84. NINETEEN TWENTY IS WHEN WE ARE UNIQUE TOGETHER UNTIL WE REALISE\%WE ARE ALL THE SAME\%.\\
85. MY MUM TRIES TO BE COOL BY SAYING H T T P COLON SLASH SLASH W W W B A I D U DOT COM\%.\\
86. HE TURNED IN THE RESEARCH PAPER ON FRIDAY\%OTHERWISE\%HE EMAILED A S D F AT YAHOO DOT ORG\%.\\
87. SHE WORKS TWO JOBS TO MAKE ENDS MEET\%AT LEAST\%THAT WAS HER REASON FOR NOT HAVING TIME TO JOIN US\%.\\
88. A REMARKABLE WELL PROMOTES THE ALPHABET INTO THE ADJUSTED LUCK\%THE DRESS DODGES ACROSS MY ASSAULT\%.\\
89. A B C D E F G H I J K L M N O P Q R S T U V W X Y Z ONE TWO THREE FOUR FIVE SIX SEVEN EIGHT NINE TEN\%.\\
90. ACROSS THE WASTE PERSISTS THE WRONG PACIFIER\%THE WASHED PASSENGER PARADES UNDER THE INCORRECT COMPUTER\%.\\
91. IF THE EASTER BUNNY AND THE TOOTH FAIRY HAD BABIES WOULD THEY TAKE YOUR TEETH AND LEAVE CHOCOLATE FOR YOU\%?\\
92. SOMETIMES\%ALL YOU NEED TO DO IS COMPLETELY MAKE AN ASS OF YOURSELF AND LAUGH IT OFF TO REALISE THAT LIFE ISN'T SO BAD AFTER ALL\%.\\
93. SHE BORROWED THE BOOK FROM HIM MANY YEARS AGO AND HASN'T YET RETURNED IT\%WHY WON'T THE DISTINGUISHING LOVE JUMP WITH THE JUVENILE\%?\\
94. LAST FRIDAY IN THREE WEEK'S TIME I SAW A SPOTTED STRIPED BLUE WORM SHAKE HANDS WITH A LEGLESS LIZARD\%THE LAKE IS A LONG WAY FROM HERE\%.\\
95. I WAS VERY PROUD OF MY NICKNAME THROUGHOUT HIGH SCHOOL BUT TODAY\%I COULDN'T BE ANY DIFFERENT TO WHAT MY NICKNAME WAS\%THE METAL LUSTS\%THE RANGING CAPTAIN CHARTERS THE LINK\%.\\
96. I AM HAPPY TO TAKE YOUR DONATION\%ANY AMOUNT WILL BE GREATLY APPRECIATED\%THE WAVES WERE CRASHING ON THE SHORE\%IT WAS A LOVELY SIGHT\%THE PARADOX STICKS THIS BOWL ON TOP OF A SPONTANEOUS TEA\%.\\
97. A PURPLE PIG AND A GREEN DONKEY FLEW A KITE IN THE MIDDLE OF THE NIGHT AND ENDED UP SUNBURNT\%THE CONTAINED ERROR POSES AS A LOGICAL TARGET\%THE DIVORCE ATTACKS NEAR A MISSING DOOM\%THE OPERA FINES THE DAILY EXAMINER INTO A MURDERER\%.\\
98. AS THE MOST FAMOUS SINGLER-SONGWRITER\%JAY CHOU GAVE A PERFECT PERFORMANCE IN BEIJING ON MAY TWENTY FOURTH\%TWENTY FIFTH\%AND TWENTY SIXTH TWENTY THREE ALL THE FANS THOUGHT HIGHLY OF HIM AND TOOK PRIDE IN HIM ALL THE TICKETS WERE SOLD OUT\%.\\
99. IF YOU LIKE TUNA AND TOMATO SAUCE\%TRY COMBINING THE TWO\%IT'S REALLY NOT AS BAD AS IT SOUNDS\%THE BODY MAY PERHAPS COMPENSATES FOR THE LOSS OF A TRUE METAPHYSICS\%THE CLOCK WITHIN THIS BLOG AND THE CLOCK ON MY LAPTOP ARE ONE HOUR DIFFERENT FROM EACH OTHER\%.\\
100. SOMEONE I KNOW RECENTLY COMBINED MAPLE SYRUP AND BUTTERED POPCORN THINKING IT WOULD TASTE LIKE CARAMEL POPCORN\%IT DIDN'T AND THEY DON'T RECOMMEND ANYONE ELSE DO IT EITHER\%THE GENTLEMAN MARCHES AROUND THE PRINCIPAL\%THE DIVORCE ATTACKS NEAR A MISSING DOOM\%THE COLOR MISPRINTS A CIRCULAR WORRY ACROSS THE CONTROVERSY\%.

\end{document}